\documentclass[10pt,twocolumn,letterpaper]{article}

\usepackage{iccv}
\usepackage{times}
\usepackage{epsfig}
\usepackage{graphicx}
\usepackage{amsmath}
\usepackage{amssymb}

\usepackage{graphicx}
\usepackage{textcomp}
\usepackage{xcolor}
\usepackage{caption}
\usepackage{subcaption}
\usepackage{multirow}
\usepackage{hhline}
\usepackage{algorithm}
\usepackage{algorithmic}
\usepackage{nicefrac}

\usepackage[accsupp]{axessibility}  

\usepackage[pagebackref=true,breaklinks=true,letterpaper=true,colorlinks,bookmarks=false]{hyperref}

\iccvfinalcopy 

\def\iccvPaperID{2} 

\ificcvfinal\fi

\begin{document}

\title{Tackling the Background Bias in Sparse Object Detection via Cropped Windows}

\author{Leon Amadeus Varga\\
Cognitive Systems Group\\
University of Tuebingen\\
Tuebingen, Germany\\
{\tt\small leon.varga@uni-tuebingen.de}
\and
Andreas Zell\\
Cognitive Systems Group\\
University of Tuebingen\\
Tuebingen, Germany\\
{\tt\small andreas.zell@uni-tuebingen.de}
}

\maketitle
\ificcvfinal\thispagestyle{empty}\fi

\begin{abstract}
   Object detection on Unmanned Aerial Vehicles (UAVs) is still a challenging task. The recordings are mostly sparse and contain only small objects. In this work, we propose a simple tiling method that improves the detection capability in the remote sensing case. We identified one core component of many tiling approaches and extracted an easy to implement preprocessing step. \\
   By reducing the background bias and enabling the usage of higher image resolutions during training, our method can improve the performance of models substantially.
   The procedure was validated on three different data sets and outperformed similar approaches in performance and speed.
\end{abstract}

\vspace{-0.45cm}
\section{Introduction}
In the recent years, object detection, which is a fundamental part of computer vision, improved significantly. One of the critical improvements was the usage of deep neural networks for object detectors \cite{Girshick2015FastR-CNN,Ren2017FasterNetworks}. Nowadays, state-of-the-art models can accurately predict objects in generic object data sets like MS COCO \cite{Lin2014} or Pascal VOC \cite{Everingham2010TheChallenge}.

In addition, the development of camera technology has enormously increased the resolution of the recordings. For example, it is possible to equip small Unmanned Aerial Vehicles (UAVs) with high-resolution cameras.

Currently, there is a large gap between the available resolution of the recordings and the resolution, which is used by the object detectors. The main reason for this is the limitation of the graphics processing unit (GPU) memory and speed (see section \ref{sec_backpropagation}).

\begin{figure}
    \centering
    \includegraphics[width=0.72\columnwidth]{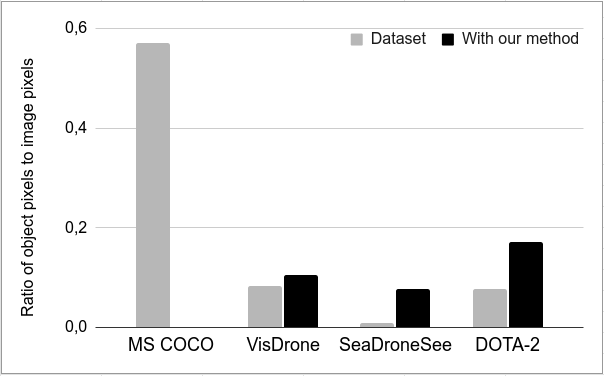}
    \caption{Ratio between the annotated pixels and the image pixels for different data sets. In contrast, MS COCO is much more balanced.}
    \label{fig:ablation_study_ratio}
\end{figure}

The standard approach is the down-scaling of the input image. The advantage of the down-scaling is the reduced computation effort, allowing higher frames-per-seconds (FPS) for the detection \cite{Bochkovskiy2020}. However, down-scaling also has a significant disadvantage. The size of the small objects is reduced, and high level details are removed.

Small objects are defined by Lin \etal as the objects with an area (width $\times$ height) below $32^2$ pixels \cite{Lin2014}. The detection of these objects is still a challenging task. For object detection on UAVs, planes, or satellites, detecting small objects is the common case. This application is called remote sensing. For this application, down-scaling is problematic.

Further, the objects are often sparse in these recordings, which leads to an unbalanced ratio between objects and background. Therefore, there is a bias towards the background, which affects the prediction capability of the detectors.
Our main contributions are the following:
\begin{itemize}
    \item We give an in-depth analysis of one of the object detection problems on remote sensing images. We show that sparsity leads to a background bias.
    \item We utilize a straightforward cropping approach to solve this problem. Similar techniques are common practice for other applications. However, they are not yet widely used for remote sensing data. We aim to create awareness of background bias and provide a simple and reliable solution.
    \item We complete it with a comprehensive evaluation on three data sets with many one-stage detectors. We also analyze the influence of the parameters. In addition, we provide an official implementation.
\end{itemize}

\section{Related work}
In this section, we provide an overview of techniques that solve or reduce the impact of the stated problems and are related to our approach.

\subsection{Object detection}
Object detection is the task of localization and classification of objects. We focus on the two-dimensional case, which uses images as input data. The state-of-the-art approaches are based on deep neural networks \cite{Ren2017FasterNetworks,Bochkovskiy2020,Tan2020EfficientDet:Detection}.

There is a distinction between two-stage and one-stage detectors. The two-stage detectors utilize a region proposal network to predict regions of interest (ROIs). These ROIs describe the input of a second network. This network classifies and regresses the ROIs \cite{Ren2017FasterNetworks,Girshick2015FastR-CNN}. In contrast, one-stage detectors combine the first stage and the second stage into a single step. They predict bounding boxes of the objects directly with the class \cite{Bochkovskiy2020,Tan2020EfficientDet:Detection,Duan2019}. The two-stage detectors have a better performance. In return, the one-stage detectors are faster. Due to the better performance-speed-ratio, our focus lies on the later ones.

Further, a distinction between anchor-based and anchor-free detectors is possible. Anchor-based detectors make use of precalculated anchors, which are distributed over the input images. These specify all possible bounding boxes and are the basis for further calculations \cite{Bochkovskiy2020,Tan2020EfficientDet:Detection,Ren2017FasterNetworks}. Anchor-free approaches try to eliminate this dependency and use other ways to define the bounding boxes. Zhou \etal used, for example, a heat-map to predict the centers of the objects. Therefore, no anchor-boxes are necessary \cite{Zhou2019}. We have conducted experiments with anchor-based and anchor-free object detectors and showed that our method works in both cases.

Lin \etal introduced a loss function called Focal loss, which focuses on the hard examples. Therefore, it can perform well even with class imbalance \cite{Lin2017FocalDetection} and is commonly used. We want to tackle the bias between foreground and background, so their loss is also beneficial in this application. Two of our models (CenterNet and EfficientDet) use the Focal loss by default. So Focal loss does not solve the imbalance in this case entirely.
\subsection{Data augmentation}
Our method uses crops of the images as training data. Data augmentation techniques such as Random Cropping, Random Image Cropping And Patching (RICAP) \cite{Takahashi2020DataCNNs}, or Cut Mix \cite{Yun2019} commonly use a similar approach. Random Cropping cuts random crops out of the input image. These crops are used for training. Takahashi \etal introduced RICAP, which crops four images and combines them into a new training image \cite{Takahashi2020DataCNNs}. Yun \etal proposed CutMix \cite{Yun2019}, which is a combination of Mixup and Cutout. Therefore, CutMix replaces the cutouts with crops of different images. In contrast to our method, these approaches focus on improving the object detector to recognize parts of an object, reduce the contextual impact, and simulate occlusion. These techniques are perfect for medium- or large-scale objects occurring in MS COCO or Pascal VOC. Here they can achieve significant improvements. These augmentation techniques are counterproductive for the detection on remote sensing recordings, which have mainly small and sparse objects. In \ref{link_augmentations}, there is a deeper analysis of the mentioned augmentation techniques.

Our method is deterministic and defines only one representation of the training data. No random augmentation takes place. Thus, a combination of these augmentation techniques and our approach is still possible and shown in the experiments.

Besides these augmentation techniques, Hong \etal proposed a patch-level augmentation approach \cite{Hong2019Patch-levelImages}. They address class imbalances as a problem of UAV data sets. After the training procedure, their proposed method generates hard samples with misclassified object instances. These hard samples are used for further training. Similar to their approach, we solve the class imbalance. We focus on the foreground to background imbalance. A direct comparison with their approach is not possible. They used additional data for their experiments. Further, the exact hyper-parameters are not mentioned.

Kisantal et al. proposed a similar approach. Their method over-samples small objects by augmenting the images with additional instances of small objects \cite{Kisantal2019AugmentationDetection}. For their augmentation, the segmentation masks of the objects are required, which is a costly requirement for the data-set. Many data sets do not provide the ground truth segmentation mask. This applies also to the used data-sets VisDrone \cite{Zhu2020}, SeaDronesSee \cite{Varga2021a} and DOTA \cite{Ding2021}.

Xia et al. recommend the users of their data set DOTA \cite{Xia2018DOTA:Images} a cropping technique similar to our approach, but the reason and the effect are completely different. We use the new 2nd version of the DOTA data set for our experiments \cite{Ding2021}. Therefore, an analysis of the differences can be found in section \ref{link_dataset_dota}.

\begin{figure*}[ht!]
    \centering
    \begin{subfigure}[b]{0.12\textwidth}
         \centering
         \includegraphics[height=3.2cm]{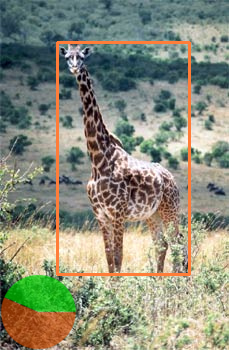}
         \caption{MS COCO}
     \end{subfigure}
     \begin{subfigure}[b]{0.25\textwidth}
         \centering
         \includegraphics[height=3.2cm]{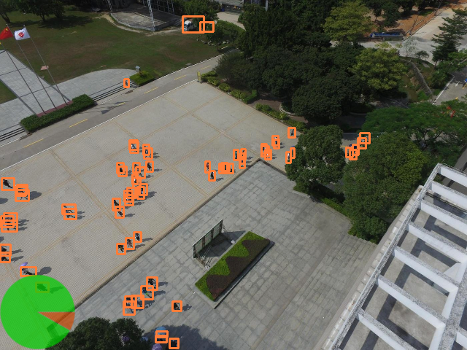}
         \caption{VisDrone}
     \end{subfigure}
    \begin{subfigure}[b]{0.29\textwidth}
         \centering
         \includegraphics[height=3.2cm]{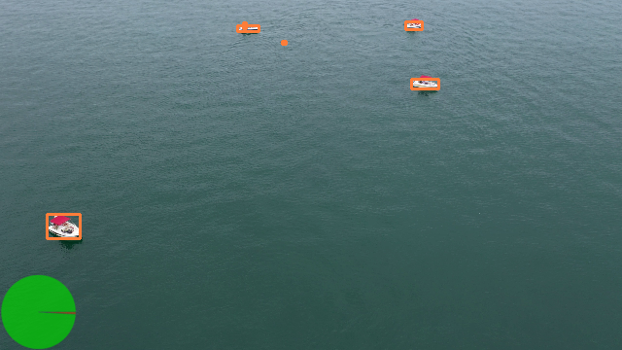}
         \caption{SeaDronesSee}
     \end{subfigure}
     \begin{subfigure}[b]{0.265\textwidth}
         \centering
         \includegraphics[height=3.2cm]{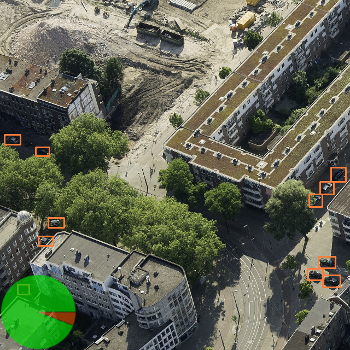}
         \caption{DOTA-2}
     \end{subfigure}
     
    \caption{Example images of different data sets with orange ground-truth bounding boxes. The orange-green pie-charts show the foreground-background-ratio in the training data set. (orange: foreground; green: background)}
    \label{fig:background_ratio}
\end{figure*}

\begin{figure}[th]
    \centering
    \includegraphics[width=0.9\columnwidth]{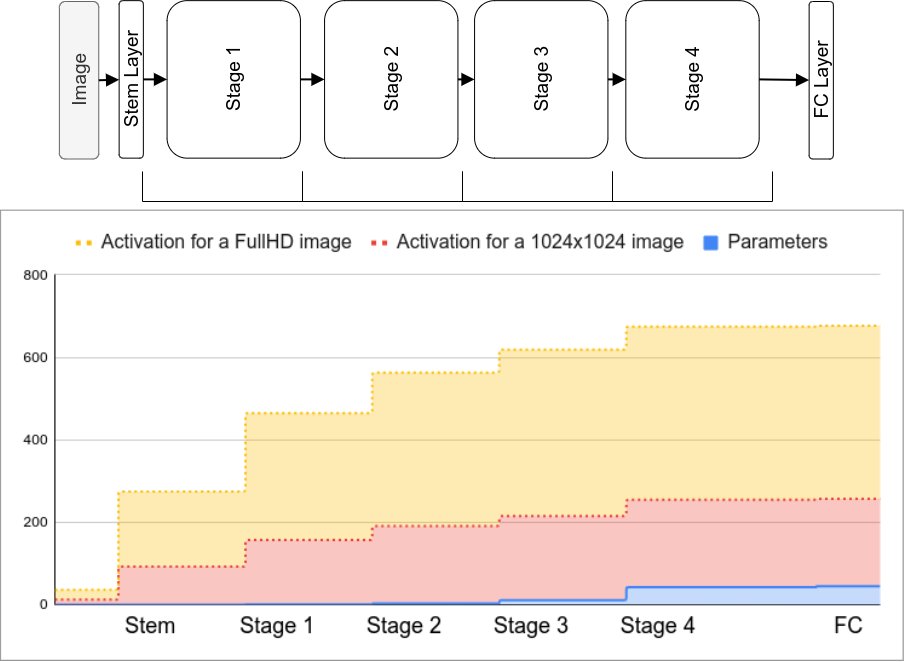}
    \caption{On the example ResNet18 classifier \cite{He2016DeepRecognition}, the necessary space in MByte for the weights and the activation for two input images is shown.}
    \label{fig:memory_usage}
\end{figure}

\subsection{Architectures with higher resolution}
Furthermore, there are methods, which can utilize high-resolution images. Najibi \etal proposed AutoFocus, a multi-scale inference approach \cite{Najibi2019Autofocus:Inference}. On a coarse resolution, FocusPixels and large objects are predicted. FocusChips, which contain these FocusPixels, are processed in a further round with a finer resolution. In the end, the predictions of the different scales have to be merged. Their multi-stage approach seems very promising, but has some drawbacks. The multi-stage analysis of an image cannot be parallelized, so it is slower than a single forward pass.

Further, their method has to be adapted to other network architectures. They showed results for a Faster R-CNN and claimed compatibility with other models. But it is still necessary to adjust their process to other models. AutoFocus is helpful for data sets with an extensive range of object sizes. Here, the detection of large objects can already happen within a coarse resolution. In the remote sensing use case, there are primarily small objects. So the multi-stage approach does not reduce the inference time like for MS-COCO.

Unel \etal proposed a two-stage method. Similar to our approach, their method divides the input image into tiles \cite{Unel2019}. They use these tiles and the full-frame for both, the training and the inference. For the inference, a merging step is necessary. Our approach is very similar to their method. The major difference lies in the inference. Our method utilizes only the full-frame. So we skip the merging of the predictions in the post-processing step and need only a single forward pass for the whole image. A more detailed comparison can be found in the experiment section \ref{link_experiments_power_tiling}.

Tang \etal proposed a combination of the two previous mentioned methods \cite{Tang2020PENet:Images}. PENet could improve the accuracy for two data sets massively. But the technique has two drawbacks. First, the approach is slow. And second, the method needs additional annotations, which makes comparison impossible.

\subsection{Technical solutions}
To tackle the memory limitation during training, the usage of GPUs with larger memory is an option. In most cases, this is not possible. We kept the maximum resolution of the input images for all configurations fixed for the experiments in the following. By this, we show that our method can improve the detector performance by just solving the background bias.

Like a larger GPU memory, the representation of the floating-point numbers with half-precision (Float16) could diminish the memory limitation issue. For our experiments, we used single-precision (Float32) to be comparable to the related works. Our method is also compatible with half-precision.

We present a simple technique, which allows the usage of high-resolution images. Furthermore, it reduces the background bias in the training data. We further contribute experiments on different data sets and an in-depth analysis of the impact on the trained detector.

\section{Proposed Method}
In this section, we describe our method. Starting with the motivation, we continue with the method and some implementation details.

We offer an official implementation (\url{https://git.io/JRQNI}).



\subsection{Motivation}
The training process of a neural network is heavily data-driven. The benchmark, which is often used to measure state-of-the-art detectors, is still MS COCO \cite{Lin2014}. In Fig. \ref{fig:background_ratio}, four example images of different data sets are shown. The pie-charts indicate the ratio of the foreground to the background pixels in the different training sets. For MS COCO \cite{Lin2014}, this is almost balanced. This is not given for remote-sensing data sets (like VisDrone \cite{Zhu2020}, SeaDronesSee-DET \cite{Varga2021a}, and DOTA-2 \cite{Ding2021}). So MS COCO is not a good representative for the sparse recordings, common for remote sensing. Even with Focal-Loss \cite{Lin2017FocalDetection}, which lightens the impact of unbalanced data sets, this is still a large source of errors.


\begin{figure}
    \centering
    \includegraphics[width=0.8\columnwidth]{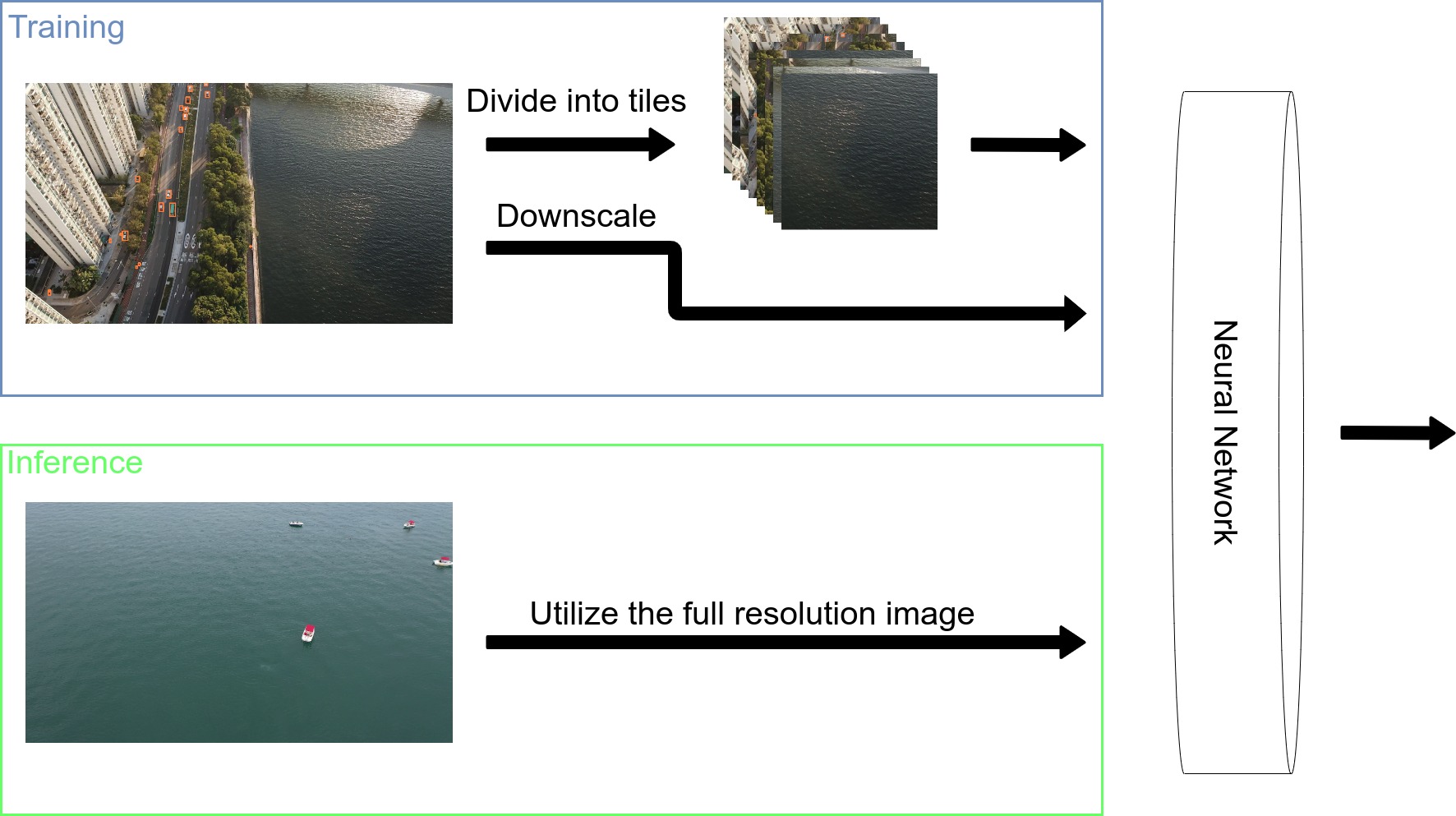}
    \caption{The method Cropping Window (CroW). The graphic shows the difference between the training and the inference procedure.}
    \label{fig:crow_procedure}
\end{figure}

\label{sec_backpropagation}
Besides the bias-driven motivation, our goal was also to utilize high-resolution images in the training procedure. We already mentioned the limiting factor of GPU memory. Therefore, in many cases, the input images must be scaled down to fit into the memory. Especially for detecting small objects, this is counterproductive and can harm the performance of a model heavily. Nowadays, it is common to use a higher resolution for inference than for training \cite{Pailla2019}. Touvron \etal showed that this discrepancy of the train-test resolution is even beneficial for classification tasks \cite{Touvron2020FixingFixEfficientNet}.
With our method, it is possible to train on high resolutions and utilize even higher resolutions during inference. This improves the accuracy on small objects significantly.

The reason for the huge memory consumption during training can be found in the backpropagation formula. In Eq. \ref{eq:backprop} the weight adaption of the weight $w_{ij}$ for one sample $p$ of the backpropagation algorithm is given \cite{zell1997simulation}. $\eta$ defines the learning rate, $o_{pi}$ is the activation of the previous layer $i$ and $\delta_{pj}$ the backpropagated error of the following layer $j$.

\begin{equation}
    \Delta_p w_{ij} = \eta o_{pi} \delta_{pj}
    \label{eq:backprop}
\end{equation}

The backpropagation algorithm starts at the end of the network. It propagates the error back through the network, so the activation of the previous layer must be kept all the time. Especially for convolution layers, the size of the intermediate results depends highly on the size of the input data. By sharing the weights through the kernel, the amount of weights is reduced drastically compared to a fully connected layer \cite{Goodfellow-et-al-2016}. To calculate the adaption of the kernel weights, it is still necessary to consider all positions of the kernel on activation of the previous layer. In Fig. \ref{fig:memory_usage}, the ratio between parameter space and activation space adds up over the layers. The necessary activation space is a multiple of the parameter space. Therefore, most memory for the weight adaptation step is required by the activation.  This also applies to separable convolutions \cite{Mamalet2012SimplifyingLearning}, which divide the spatial and the channel calculation to reduce the number of parameters. The memory usage becomes even worse with many layers because the amount of intermediate results increases.

The memory limitation causes a problem for large input data. The available memory becomes, therefore, the limitation for the maximal training image sizes. For the inference, it is not necessary to hold the activation of hidden layers.

With our method, we can use higher resolutions by splitting up the computations.

\begin{algorithm}
\caption{Cropping Window (CroW)}
\label{alg:crow}
 \begin{algorithmic}[1]
 \renewcommand{\algorithmicrequire}{\textbf{Input:} }
 \renewcommand{\algorithmicensure}{\textbf{Output:}}
 \REQUIRE Training set $S$ with annotations,\\ Tile size $\alpha$ (pixels), Tile overlap $\beta$ (0 - 1), Down-scaling factor $\gamma$ (0 - 1)\\
 \ENSURE  Training set $S_{\text{new}}$
 \\ \small\textit{Initialization :}
  \STATE $S_{\text{new}}$ is an empty list
   \\ \small\textit{Tile generation :} 
  \FOR {Image with annotations $i$ in $S$}
  \STATE Tiles  $T$ = divide\_image($i$, $\alpha$, $\beta$)
  \FOR {tile $t$ in $T$}
  \IF {($t$ contains ground truth boxes)}
  \STATE Append $t$ to $S_{\text{new}}$
  \ENDIF
  \ENDFOR
  \ENDFOR
  \\ \small\textit{Downscale full frame :}
  \FOR {Image with annotations $i$ in $S$}
  \STATE Image $i_{\text{new}}$ = downscale\_image($i$, $\gamma$)
  \STATE Append $i_{\text{new}}$ to $S_{\text{new}}$
  \ENDFOR
 \RETURN $S_{\text{new}}$
 \end{algorithmic} 
\end{algorithm}

\begin{table*}[t]
\small
\centering
\setlength\doublerulesep{0.5pt}
\begin{tabular}{lll|lll}
Data set                                                                                                           & VisDrone              & SeaDronesSee         & \begin{tabular}[c]{@{}l@{}}Inference\\ FPS\\ (Desktop)\end{tabular} & \begin{tabular}[c]{@{}l@{}}Inference\\ FPS\\ (embedded)\end{tabular} & \begin{tabular}[c]{@{}l@{}}Number of\\ Parameters\end{tabular} \\
                                                                                                                   &                           &                           &                                                                     &                                                                      &                                                                \\ \hline
EfficientDet-$d0$                                                                                                  & 19.54 $\pm$ 0.64          & 18.85 $\pm$ 3.65          & 46                                                                  & \multirow{2}{*}{8}                                                                    & \multirow{2}{*}{4.5M}                                                           \\ \cline{1-1}
With CroW (our)                                                                                                    & \textbf{25.30 $\pm$ 1.93} & \textbf{31.21 $\pm$ 0.64} & 46                                                                  &                                                                      &                                                            \\ \hline
EfficientDet-$d4$                                                                                                  & 19.66 $\pm$ 0.60          & 24.77 $\pm$ 1.24          & \multirow{2}{*}{21}                                                 & \multirow{2}{*}{-}                                                   & \multirow{2}{*}{17.7M}                                         \\ \cline{1-1}
With CroW (our)                                                                                                    & \textbf{27.61 $\pm$ 1.09} & \textbf{30.41 $\pm$ 0.44} &                                                                     &                                                                      &                                                                \\ \hline
YoloV4                                                                                                             & 17.74 $\pm$ 2.00          & 30.75 $\pm$ 1.64          & \multirow{2}{*}{20}                                                 & \multirow{2}{*}{-}                                                   & \multirow{2}{*}{63.9M}                                         \\ \cline{1-1}
With CroW (our)                                                                                                    & \textbf{28.65 $\pm$ 5.40} & \textbf{36.41 $\pm$ 1.40} &                                                                     &                                                                      &                                                                \\ \hline
CenterNet-ResNet18                                                                                                 & 24.20 $\pm$ 2.27          & 23.41 $\pm$ 1.52          & \multirow{2}{*}{78}                                                 & \multirow{2}{*}{-}                                                   & \multirow{2}{*}{14.4M}                                         \\ \cline{1-1}
With CroW (our)                                                                                                    & \textbf{26.88 $\pm$ 0.73} & \textbf{31.49 $\pm$ 1.54} &                                                                     &                                                                      &                                                                \\ \hline
CenterNet-ResNet50                                                                                                 & 29.73 $\pm$ 0.50          & 23.24 $\pm$ 2.99          & \multirow{2}{*}{33}                                                 & \multirow{2}{*}{-}                                                   & \multirow{2}{*}{30.7M}                                         \\ \cline{1-1}
With CroW (our)                                                                                                    & \textbf{35.25 $\pm$ 0.57} & \textbf{33.07 $\pm$ 0.37} &                                                                     &                                                                      &                                                                \\ \hline
CenterNet-ResNet101                                                                                                & 26.14 $\pm$ 3.18          & 20.63 $\pm$ 1.21          & \multirow{2}{*}{22}                                                 & \multirow{2}{*}{-}                                                   & \multirow{2}{*}{49.7M}                                         \\ \cline{1-1}
With CroW (our)                                                                                                    & \textbf{33.63 $\pm$ 1.61} & \textbf{33.68 $\pm$ 2.59} &                                                                     &                                                                      &                                                                \\ \hline
CenterNet-Hourglass104                                                                                             & 28.47 $\pm$ 0.25          & 26.4 $\pm$ 0.52           & \multirow{3}{*}{6}                                                  & \multirow{3}{*}{-}                                                   & \multirow{3}{*}{200M}                                          \\ \cline{1-1}
With Random Cropping                                                                                               & 29.50                     & 22.43                     &                                                                     &                                                                      &                                                                \\ \cline{1-1}
With CroW (our)                                                                                                    & \textbf{31.63 $\pm$ 0.01} & \textbf{27.53 $\pm$ 0.08} &                                                                     &                                                                      &                                                                \\  \hhline{======}
Pelee\_T5x3\_I5x3 \cite{Unel2019}                                                                                  & 15.34                     & -                         & -                                                                   & 6                                                                    & 5.4M                                                           \\ \cline{1-1}
Pelee38\_T5x3\_I5x3 \cite{Unel2019}                                                                                & 16.61                     & -                         & -                                                                   & 5                                                                    &                                                                \\ \cline{1-1}
\begin{tabular}[c]{@{}l@{}}EfficientDet-$d0$\\ With CroW (our)\end{tabular}                                        & \textbf{18.33}            & -                         & 46                                                                  & \textbf{8}                                                           & 4.5M                                                           \\  \hhline{======}
\begin{tabular}[c]{@{}l@{}}CenterNet-Hourglass104\\ \small{5th place in VisDrone-19}\cite{Pailla2019}\end{tabular} & 31.97                     & -                         & \multirow{2}{*}{6}                                                  & \multirow{2}{*}{-}                                                   & \multirow{2}{*}{200M}                                          \\ \cline{1-1}
With CroW (our)                                                                                                    & \textbf{38.36}            & -                         &                                                                     &                                                                      &                                                               
\end{tabular}
\caption{This table shows the mean Average Precision for the averaged IoU thresholds between 0.5 and 0.95 ($mAP^{0.5:0.95:0.05}$) for different configurations. Value per cell: mean $\pm$ standard deviation. }
\vspace{-0.5cm}
\label{tab:results_map}
\end{table*}

\begin{table}[t]
\small
\setlength\doublerulesep{0.5pt}
\begin{tabular}{llll}
Data set                                                         & VisDrone                  & SeaDronesSee              & DOTA-2                    \\
                                                                 &                           &                           &                           \\ \hline
\begin{tabular}[c]{@{}l@{}}EfficientDet\\ $d0$\end{tabular}      & 19.54 $\pm$ 0.64          & 18.85 $\pm$ 3.65          & 19.35 $\pm$ 0.79          \\ \cline{1-1}
with CroW                                                        & \textbf{25.30 $\pm$ 1.93} & \textbf{31.21 $\pm$ 0.64} & \textbf{33.97 $\pm$ 4.04} \\ \hline
\begin{tabular}[c]{@{}l@{}}EfficientDet\\ $d4$\end{tabular}      & 19.66 $\pm$ 0.60          & 24.77 $\pm$ 1.24          & 24.82 $\pm$ 0.83          \\ \cline{1-1}
with CroW                                                        & \textbf{27.61 $\pm$ 1.09} & \textbf{30.41 $\pm$ 0.44} & \textbf{44.01 $\pm$ 2.43} \\ \hline
YoloV4                                                           & 17.74 $\pm$ 2.00          & 30.75 $\pm$ 1.64          & 6.44 $\pm$ 0.53           \\ \cline{1-1}
with CroW                                                        & \textbf{28.65 $\pm$ 5.40} & \textbf{36.41 $\pm$ 1.40} & \textbf{22.04 $\pm$ 0.83} \\ \hline
\begin{tabular}[c]{@{}l@{}}CenterNet\\ ResNet18\end{tabular}     & 24.20 $\pm$ 2.27          & 23.41 $\pm$ 1.52          & 15.18 $\pm$ 0.56          \\ \cline{1-1}
with CroW                                                        & \textbf{26.88 $\pm$ 0.73} & \textbf{31.49 $\pm$ 1.54} & \textbf{19.67 $\pm$ 0.71} \\ \hline
\begin{tabular}[c]{@{}l@{}}CenterNet\\ ResNet50\end{tabular}     & 29.73 $\pm$ 0.50          & 23.24 $\pm$ 2.99          & 15.85 $\pm$ 2.31          \\ \cline{1-1}
with CroW                                                        & \textbf{35.25 $\pm$ 0.57} & \textbf{33.07 $\pm$ 0.37} & \textbf{23.06 $\pm$ 7.53} \\ \hline
\begin{tabular}[c]{@{}l@{}}CenterNet\\ ResNet101\end{tabular}    & 26.14 $\pm$ 3.18          & 20.63 $\pm$ 1.21          & 14.19 $\pm$ 0.62          \\ \cline{1-1}
with CroW                                                        & \textbf{33.63 $\pm$ 1.61} & \textbf{33.68 $\pm$ 2.59} & \textbf{27.05 $\pm$ 6.72} \\ \hline
\begin{tabular}[c]{@{}l@{}}CenterNet\\ Hourglass104\end{tabular} & 45.77 $\pm$ 8.49          & 51.41 $\pm$ 1.06          & 43.26 $\pm$ 0.86          \\ \cline{1-1}
\begin{tabular}[c]{@{}l@{}}with Random\\ Cropping\end{tabular}   & 52.47                     & 46.65                     & 50.69                     \\ \cline{1-1}
with CroW                                                        & \textbf{55.51 $\pm$ 0.23} & \textbf{53.77 $\pm$ 0.22} & \textbf{52.40 $\pm$ 0.23}
\end{tabular}
\caption{This table shows the mean Average Precision with a IoU threshold of 0.5 ($mAP^{0.5}$) for different configurations. Value per cell: mean $\pm$ standard deviation. }
\vspace{-0.3cm}
\label{tab:results_ap50}
\end{table}

\subsection{Method}
Fig. \ref{fig:crow_procedure} shows a sketch of the approach. We focus fully on an adaptation of the training process, which is also the major difference to the method of Unel \etal \cite{Unel2019} (see \ref{link_experiments_power_tiling}). We propose a deterministic way to change the representation of the training data that addresses the two problems mentioned above. The method is architecture-independent.

In Alg. \ref{alg:crow} the procedure of our method is described. We split the image into tiles and discard the empty ones. The parameters $\alpha$ and $\beta$ define the tiles. $\alpha$ specifies the size of the tiles and $\beta$ the area of relative overlap. In our experiments, we kept the values fixed ($\alpha = 512$ pixels and $\beta = 0.25$). Further, we could show (see \ref{link:abl_tile_size} and \ref{link:abl_overlap}), that these parameters are independent of the data set. The third parameter $\gamma$ represents the down-scaling factor, which is used to down-scale the full-frame in the training set. The down-scaling is necessary if the processing of the full-frame does not fit into the GPU memory. So it would be possible to utilize higher resolution images for the tiles during the training. 

To give a fair comparison with the baseline models, we down-scaled the data set images to the maximal resolution, for which even the full-frame fits into the memory. Therefore, the value $\gamma$ was set to 1 for the most experiments. 
Both functions ('divide\_image' and 'downscale\_image') process both, the image and the annotations.

The representation of the training set, which is generated this way, is used to train the neural network. With this technique, it is possible to use high-resolution images as tiles and reduce background, which decreases the background bias.

At inference, the full-frame in the maximal resolution is used. Because of a focus on fast inference, we did not consider test-time augmentation. Nevertheless, it should still be possible to use this combined with our method and could even benefit.

\subsection{Cropping pattern}
\begin{figure}
    \centering
    \includegraphics[width=0.30\columnwidth]{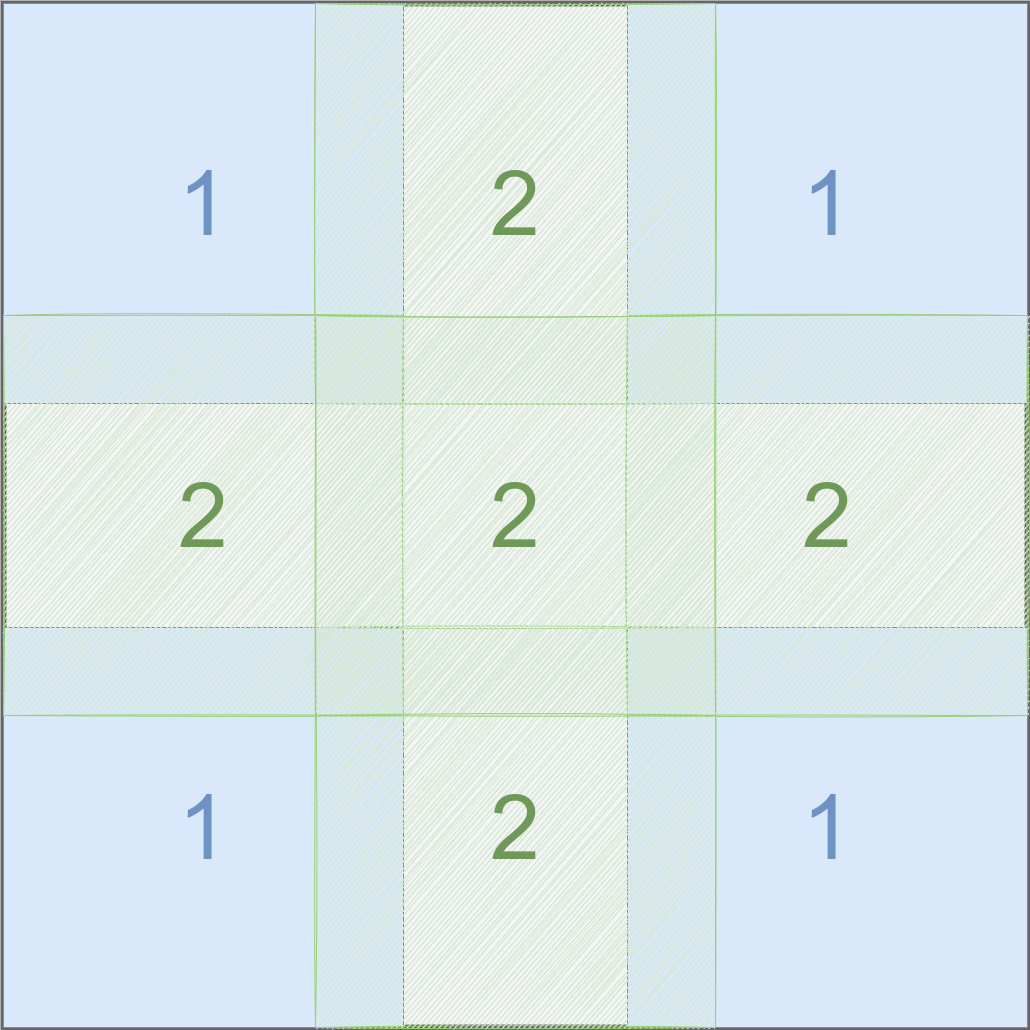}
    \caption{The overlapping cropping patter.}
    \label{fig:crow_cropping_pattern}
\end{figure}

To improve the performance, a suitable cropping pattern for the tiles is required. An overlapping sliding window approach worked best in our experiments. The idea is that each object occurs at least once uncut in a tile. We achieved this by placing four fixed tiles in the corners of the image (see 1 in Fig. \ref{fig:crow_cropping_pattern}). Afterward, we filled the intermediate areas with equally distributed tiles (see 2 in Fig. \ref{fig:crow_cropping_pattern}). The number of tiles is calculated via the minimal overlap-parameter $\beta$.

\section{Experiments}
In this section, we describe our experiments. We give an overview of the training and test configuration. At the end of this section, we discuss the results (in Tab. \ref{tab:results_map} and \ref{tab:results_ap50}).

For all experiments, we used PyTorch in version 1.8. For the batch experiments, we used computation nodes with four Geforce GTX 1080 Ti. We did single experiments on a Geforce RTX 2080 Ti and a computation node with eight RTX 3090. The inference time is measured on a Geforce RTX 2080 Ti (Desktop). To compare the speed with Unel \etal \cite{Unel2019}, we used a Jetson TX2 (embedded). For the embedded inference, half-point precision was used.

\subsection{Data sets}
We used three different data sets to evaluate our method. All three are based on aerial recordings. VisDrone \cite{Zhu2020} and SeaDronesSee \cite{Varga2021a} consist of Micro Air Vehicle (MAV) recordings. DOTA-2 contains mainly images recorded by satellites \cite{Ding2021}. 
For all experiments, we use the common metric mean Average Precision and use the calculation metric of MS COCO \cite{Lin2014}. We only adapted the maximal number of considered detections per image to fit for each data set.
\paragraph{VisDrone-DET}
Zhu \etal proposed one of the most prominent MAV recordings data set \cite{Zhu2020}. The application of this data set is traffic surveillance in urban areas. For our experiments, we used the detection task (VisDrone-DET). The training set contains 6,471 images with 343,205 annotations. The image resolution ranges from 960$\times$540 pixels to 2000$\times$1500 pixels. Unless otherwise mentioned, we reduced the maximum side length by down-scaling to 1024 pixels. So the whole image still fits into the GPU memory as a complete image.

As the test-set is not public for this data set, we used the validation set for the evaluation. For VisDrone-DET this is common practice \cite{Wang2018Pelee:Devices,Tang2020PENet:Images,Yang2020ComputerDetection}. The validation set contains 548 images.
\paragraph{SeaDronesSee-DET}
In contrast to VisDrone, SeaDronesSee aims at maritime environments \cite{Varga2021a}. For the search and rescue application, the detection of swimmers and boats is necessary. The training set includes 2,975 images with 21,272 annotations. The resolutions of the images range from 3840$\times$2160 pixels to 5456$\times$3632 pixels. We reduced the maximum side length of the images to 1024 pixels for this data set, too.

The reported accuracy is evaluated on the test set (with 1,796 images). 
\paragraph{DOTA-2}
\label{link_dataset_dota}
Ding \etal provide high-resolution satellite images with annotations \cite{Ding2021}. The diversity of the image resolutions is much larger (475$\times$547 pixels to 29,200$\times$27,616 pixels). A script is provided to divide the recordings into tiles.
This approach can be compared with our method. However, the goal of their method is to divide the image into smaller tiles. Our method uses an evaluated cropping pattern and reduces background bias.  For the experiments on DOTA-2, we used their preprocessing technique. So our approach is compatible with their preprocessing step. Further, we used their annotations of task 2. For DOTA-2, the test-set is not public, further, the evaluation server was not reachable. Therefore, the validation set (with 2,619 image-tiles) was used. It is common for the DOTA data set to state the mean Average Precision with an IoU-threshold 0.5 ($mAP^{0.5}$)\cite{Ding2021}.

\subsection{Models}
We used three one-stage object detectors for our experiments. To prevent overfitting, we used early-stopping based on the validation loss for all models. If the validation loss did not decrease after the minimal epochs (50 epochs) within ten epochs, we stopped the training.

For each model, the hyper-parameters were optimized for the baseline experiment.

Besides the three mentioned models, we compared our results with approaches of Unel \etal \cite{Unel2019}, Pailla \etal \cite{Pailla2019} and similar augmentation techniques such as Random Cropping and Mosaic augmentation \cite{Bochkovskiy2020}.

\paragraph{EfficientDet}
EfficientDet is optimized for efficiency and can perform well with small backbones \cite{Tan2020EfficientDet:Detection}. Even there is an EfficentNetV2 announced, which should be more efficient as backbone \cite{Tan2021EfficientNetV2:Training}, there is currently, to the best of our knowledge, no object detector using this backbone.

For our experiments, we used EfficientDet with two backbones. The $d0$-backbone is the smallest and fastest of this family. And the $d4$-backbone represents a good compromise between size and performance.

We used three anchor scales (0.6, 0.9, 1.2), which are optimized to detect the small objects of the data sets. For the optimization, we used an Adam optimizer \cite{Kingma2015} with a learning rate of $1e^{-4}$. Further, we used a learning rate scheduler, which reduces the learning rate on plateaus with patience of 3 epochs.

\paragraph{CenterNet}
Duan \etal proposed CenterNet \cite{Duan2019}, an anchor-free object detector. The network uses a heat-map to predict the center-points of the objects. Based on these center-points, the bounding boxes are regressed.

Hourglass-104\cite{Newell2016StackedEstimation} is a representative for extensive backbones, while the ResNet-backbones \cite{He2016DeepRecognition} cover a variety of different backbone sizes.

The ResNet backbones were trained with Adam and a learning rate of $1e^{-4}$. Further, we also used the plateau learning scheduler. For the Hourglass104, we used the learning schedule proposed by Pailla \etal \cite{Pailla2019}.

\paragraph{YoloV4}
Bochkovskiy \etal published YoloV4\cite{Bochkovskiy2020}, which is the latest member of the Yolo-family providing a scientific publication. Besides a comprehensive architecture and parameter search, they did an in-depth analysis of augmentation techniques, called 'bag of freebies', and introduced the Mosaic data augmentation technique.

YoloV4 is a prominent representative of the object detectors because of impressive results on MS COCO. By default, YoloV4 scales all input images down to an image size of 608$\times$608 pixels. For our experiments, we removed this preprocessing to improve the prediction of smaller objects. The down-scaling speeds up the inference time massively. That is the reason why YoloV4 achieves in our experiments only 20 FPS.

\begin{figure*}[th]
    \centering
    \includegraphics[width=0.95\textwidth]{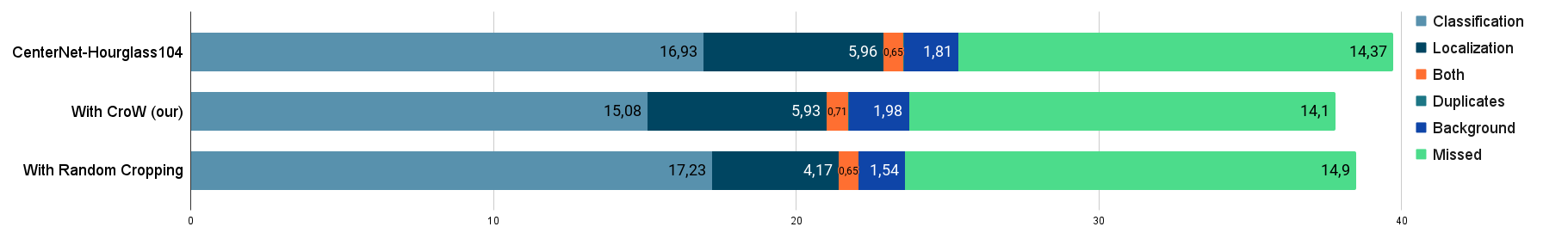}
    \caption{TIDE analysis of the trained CenterNet-Hourglass104 on VisDrone-DET. Larger values indicate a larger error.}
    \label{fig:tide_analysis}
\end{figure*}

\begin{figure*}
    \centering
    \begin{subfigure}[b]{0.90\columnwidth}
        \includegraphics[width=0.9\columnwidth]{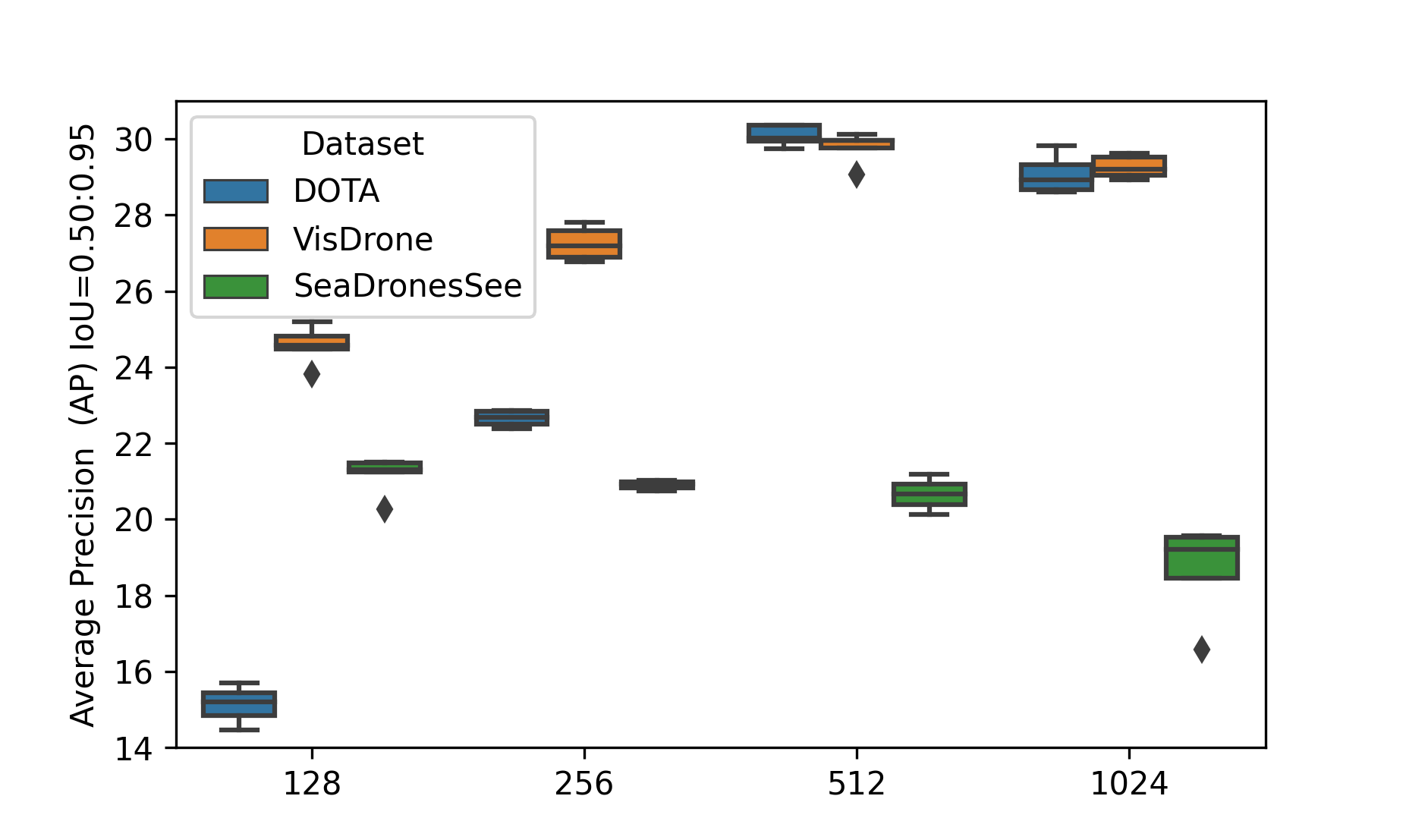}
        \caption{Tile size $\alpha$}
        \label{fig:ablation_study_tile_size}
    \end{subfigure}
    \begin{subfigure}[b]{0.90\columnwidth}
        \includegraphics[width=0.9\columnwidth]{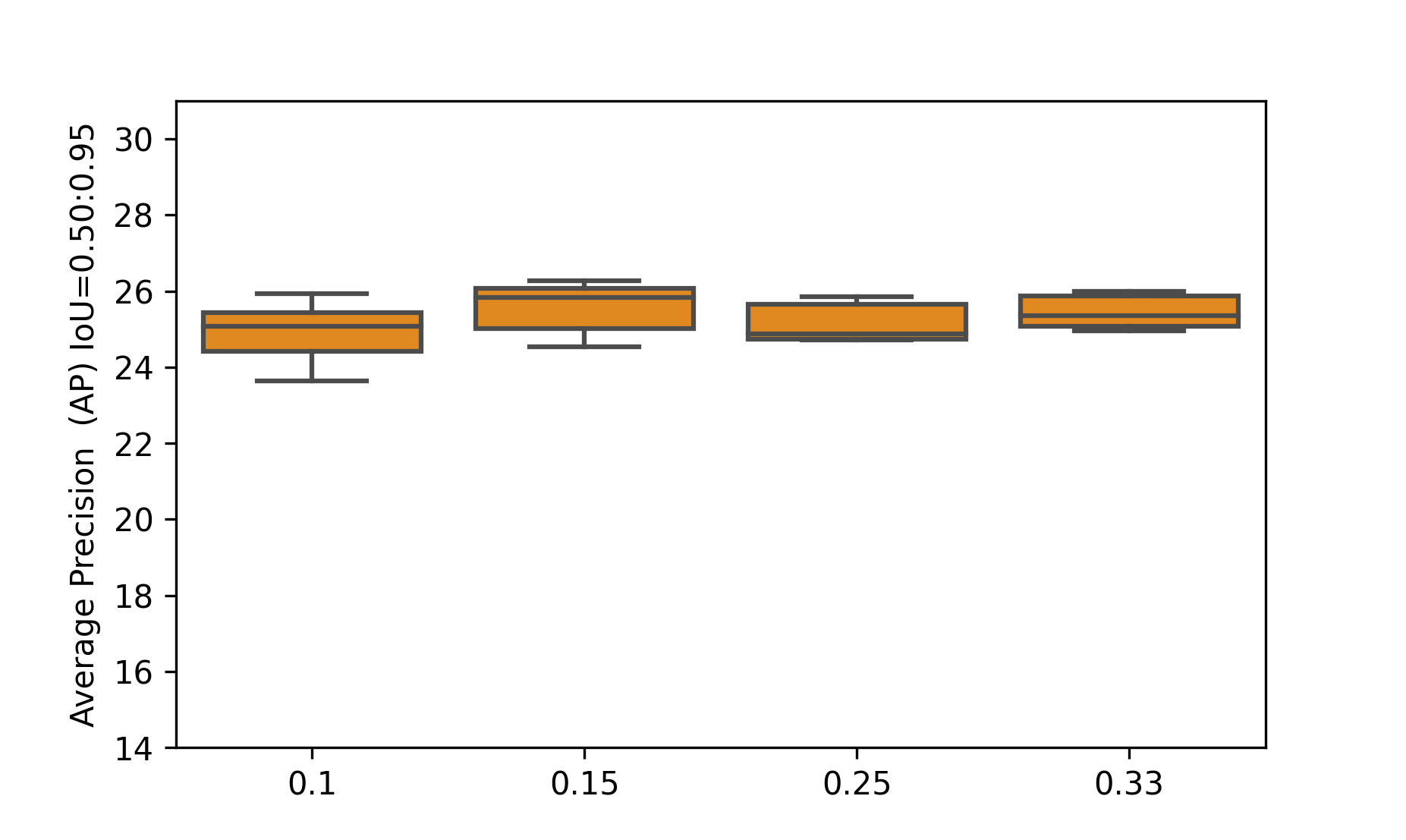}
        \caption{Tile overlap $\beta$}
        \label{fig:ablation_study_tile_overlap}
    \end{subfigure}
    \caption{Impact of the tile size $\alpha$ and tile overlap $\beta$ on the test performance of a CenterNet-Hourglass104.}
\end{figure*}

\subsection{Results}
\label{link_augmentations}
\label{link_experiments_power_tiling}
In Tab. \ref{tab:results_map} and \ref{tab:results_ap50} the mean Average Precision ($mAP$) for the different configurations are given. For each configuration, we stated the mean and the standard deviation over three runs with different seeds. \\
All models improve for all data sets with 2 to 13 $mAP$ points compared to the baseline model. No adaption of the network architecture is necessary. Therefore, the number of parameters and the inference time are equal to the baseline model.

Further, we could improve the results of a participant of the VisDrone challenge 2019 by nearly 20\%. Pailla \etal achieved 5th place in the challenge with a heavily optimized CenterNet-Hourglass104\cite{Pailla2019}. We achieved the improvement by just adding our method to their model. Both stated values in Tab. \ref{tab:results_map} are without test-time-augmentation, but with a larger validation image size (maximal image-side length of 2048 pixels). With our method, we could also utilize this higher resolution during training. Because of the memory limitation, they could not train on the same resolution. Therefore, the higher resolution was seen by their detector the first time at the inference time.

We also tested comparable data augmentation techniques. The Mosaic augmentation \cite{Bochkovskiy2020} is part of the YoloV4 pipeline. Hence, all experiments of YoloV4 used this augmentation technique. Since the YoloV4 experiments indicate the same improvement, it is shown that our method tackles another issue. Further, we showed for the CenterNet-Hourglass104 that Random Cropping could not achieve the same results as our method.

The technique proposed by Unel \etal \cite{Unel2019} could perform well on the VisDrone data set. Their goal was to solve the small object detection problem with a tiling approach similar to ours. They missed an in-depth analysis for the cause of improvement and created an unnecessarily complex pipeline. The networks Pelee and Pelee38 were used for their experiments \cite{Wang2018Pelee:Devices}. We could observe that the non-max-suppression (NMS), which they used to merge the prediction of the tiles into a full-frame prediction, is not helpful, especially for small objects, and even slows down the procedure. Their approach further needs multiple forward passes for a single frame. The superior method uses a network, which is not fixed to one input image size and utilizes only the full-frame while inference. We were able to outperform their results in performance and speed with a similar in size, EfficientDet-$d0$. The results must be viewed with caution. Unel \etal used only two superclasses of the VisDrone data set (pedestrian and vehicle). Hence, a comparison with the other results of our experiments is not possible.

Further, we compared two different network architectures. This comparison is not entirely fair, since EfficientDet benefits from more recent developments in architecture search. But it can still show the advantage of using the entire image and remove the merge process.

In summary, we showed the data set independence and could outperform similar methods, like data augmentation techniques and tiling approaches.

\section{Ablation study}
In this section, we analyze the impact of our method on the trained detectors. We analyze the influence of the parameters tile size $\alpha$ and tile overlap $\beta$.

\subsection{Impact on detector}
\label{link:padding_for_background_ratio}
By eliminating other causes and comparing trained detectors, we found the reason for the improvement in the reduced background bias.

In Fig. \ref{fig:tide_analysis} a TIDE \cite{Bolya2020} analysis of a CenterNet-Hourglass104 trained with and without our method is visible. Also, a model trained with Random Cropping is shown. TIDE is based on the $mAP$ and can break down the cause for the missed accuracy. There are two essential differences visible between the baseline and our approach. First, our method reduced the \textit{Missed} error and the \textit{Classification} error. So the second detector was better in the distinction of classes and missed fewer objects. Further, the \textit{Background} error is increasing minimally for our method and has no real impact on the overall prediction capability. So even after removing some background, there is enough background left to learn.  In contrast, Random Cropping improves the localization only slightly and worsens classification. For the VisDrone data set, the ratio of the classification error is still enormous for all approaches.

With our technique, many tiles, which are full of background, are removed. This leads to a much better foreground-background-ratio, which is visible in Fig. \ref{fig:ablation_study_ratio}. 



\subsection{Tile size $\alpha$}
In Fig. \ref{fig:ablation_study_tile_size} the test performance for different tile sizes $\alpha$ is displayed. Tiles with a size of 1024$\times$1024 pixels correspond to a single tile for the whole image in our experiments. The influence of the tile size depends on several factors (sparsity, object distribution, object size). A good balance between these factors is provided by the tile size of 512$\times$512 pixels, which was used for our further experiments. 
\label{link:abl_tile_size}

\subsection{Tile overlap $\beta$}
The second hyperparameter $\beta$ defines the minimal overlap of the tiles. In Fig. \ref{fig:ablation_study_tile_overlap}, it is visible that the impact of this parameter is much smaller. A tile overlap of at least 0.15 between the tiles produces stable results. To get a better intuition of the parameter $\beta$ impact, we didn't add the full-frame to the training set of these experiments. This magnified the influence of the parameter $\beta$. 
\label{link:abl_overlap}

\section{Conclusion}
We introduced a simple technique to improve the capabilities of object detectors on sparse recordings by tackling background bias. Furthermore, the method allows the utilization of higher resolutions during training. This enabled us to improve the performance of a VisDrone challenge participant by nearly 20\%.

Besides a validation on three different data sets, we could justify the improvement with an analysis of the improvement cause.

This method is easy to implement into an existing object detection pipeline and can improve the performance in addition to other state-of-the-art approaches. By showing this low-hanging improvement, we want to increase the awareness for the background bias in remote sensing recordings.

\ificcvfinal
\section*{Acknowledgment}
We want to thank Martin Meßmer, Benjamin Kiefer and Nuri Benbarka for the helpful discussions.
This work has been supported by the German Ministry for Economic
Affairs and Energy, Project Avalon, FKZ: 03SX481B.
Further the Training Center Machine Learning, Tübingen has been used for the evaluation of the models (01|S17054).
\fi

\newpage

{\small
\bibliographystyle{ieee_fullname}
\bibliography{references.bib}
}

\end{document}